\documentclass[conference]{IEEEtran}
\IEEEoverridecommandlockouts
\usepackage{cite}
\usepackage{amsmath,amssymb,amsfonts}
\usepackage{algorithmic}
\usepackage{graphicx}
\usepackage{textcomp}
\usepackage{xcolor}
\usepackage{svg}
\usepackage{multirow} 
\usepackage{booktabs}
\def\BibTeX{{\rm B\kern-.05em{\sc i\kern-.025em b}\kern-.08em
    T\kern-.1667em\lower.7ex\hbox{E}\kern-.125emX}}

\begin{document}
\bibliographystyle{unsrt}

\title{Cross-Session  Decoding of Neural Spiking Data via Task-Conditioned Latent Alignment}


\author{
Canyang Zhao\textsuperscript{1},
Bolin Peng\textsuperscript{1},
J. Patrick Mayo\textsuperscript{2},
Ce Ju\textsuperscript{3},
and Bing Liu\textsuperscript{1}
\thanks{
This research was supported by National Institutes of Health (NEI) awards R01EY035673 and P30EY08098 \& the Whitehall Foundation (Author: J.P. Mayo), and by the State Key Laboratory of Brain Cognition and Brain-inspired Intelligence Technology, Chinese Academy of Sciences, Beijing, China. Corresponding Authors: Ce Ju and Bing Liu.

\textsuperscript{1}Canyang Zhao, Bolin Peng, and Bing Liu are with the Institute of Automation, Chinese Academy of Sciences, Beijing, China (e-mail: zhaocanyang2024@ia.ac.cn, b70ivor@gmail.com, bing.liu@ia.ac.cn).  }%
\thanks{\textsuperscript{2}J. Patrick Mayo is with the Departments of Ophthalmology and Bioengineering, University of Pittsburgh, Pittsburgh, USA (e-mail: mayojp@pitt.edu).}%
\thanks{\textsuperscript{3}Ce Ju is with Inria, CEA, Université Paris-Saclay, Palaiseau, France (e-mail: ce.ju@inria.fr). }%
}


\maketitle

\begin{abstract}
Training a high-performing neural decoder can be difficult when only limited data are available from a recording session.
To address this challenge, we propose a Task-Conditioned Latent Alignment framework (TCLA) for cross-session neural decoding with limited target-session data. 
Building upon an autoencoder architecture, TCLA first learns a low-dimensional neural representation from a source session with sufficient data.
For target sessions with limited data,
TCLA then aligns the target latent representations to the source session in a task-conditioned manner, enabling effective transfer of learned neural representations to support decoder training in the target session.
We evaluate TCLA on the macaque motor and oculomotor center-out datasets.
Compared to baseline methods trained solely on target-session data, TCLA consistently improves decoding performance across datasets and decoding settings, with gains in the coefficient of determination of up to 0.386 for y coordinate velocity decoding in a motor dataset.
These results suggest that TCLA provides an effective strategy for transferring knowledge from source to target sessions, improving neural decoding performance under conditions with limited target-session data.
\end{abstract}


\section{INTRODUCTION}


Training a reliable neural decoder typically requires a sufficient amount of labeled recording data.
However, in practice, some recording sessions may contain only limited data\cite{wen2023rapid}, making it difficult to train a high-performing decoder from that session alone.
The key question is whether information learned from a source session with sufficient data can be transferred to improve decoder performance in a target session with limited data.

Prior work has shown that population neural activity can exhibit relatively stable low-dimensional manifolds over extended periods\cite{gallego2017neural}, motivating approaches that align low-dimensional neural manifolds across sessions.
In this context, latent variable models have shown strong potential for capturing low-dimensional neural representations and improving decoding performance\cite{pandarinath2018inferring},\cite{kapoor2024latent},\cite{ge2026energy}. 
By learning a robust low-dimensional structure in neural population activities, these models may provide useful prior knowledge for decoder training when target-session data are scarce. Furthermore, several recent studies have sought to mitigate cross-session changes in neural signals by aligning latent spaces across sessions\cite{jude2022robust}, \cite{degenhart2020stabilization},\cite{karpowicz2025stabilizing}. 
While promising, these methods typically treat neural activity as a single within-session distribution, even though task conditions can induce distinct, condition-specific manifolds,
such as those associated with different movement directions\cite{gallego2017neural}, \cite{sadtler2014neural}.

To address these limitations, our primary goal is to leverage data from a source session to improve decoding performance in a target session with limited data, by preserving the inherent task-dependent latent structure in neural activity across sessions.
To this end, we propose TCLA, the task-conditioned latent alignment framework. 
Building upon the autoencoder architecture introduced in LDNS\cite{kapoor2024latent}, TCLA learns a low-dimensional neural representation from the source session with sufficient data, and aligns neural activity from target sessions into this latent space through a task-conditioned alignment process. 


We evaluate TCLA on three neural spiking datasets collected from non-human primates performing center-out tasks, encompassing both arm movements and eye movements.
Experimental results demonstrate that the task-conditioned latent alignment framework effectively reduces cross-session variability, thus enabling improved neural decoding performance under limited-data conditions.

\section{METHODS}

Properly leveraging similarities among task conditions is essential to discovering invariant neural components across recording sessions. TCLA is thus designed as an end-to-end framework that performs task-conditioned session alignment in the latent space.
We focus on multi-session neural recordings with explicit task conditions (i.e., datasets recorded during motor or oculomotor center-out tasks with labeled behavioral conditions). Specifically, for a dataset $U$ with $M$ sessions recorded on different days, indexed by $m=1, ..., M$, each session $m$ comprises $n$ 
trials of temporally aligned neural-behavioral data.
For the $i$-th trial in session $m$, we denote the data as $\left\{ {x_m^{\left( i \right)},y_m^{\left( i \right)},l_m^{\left( i \right)}} \right\}$, where

\begin{figure*}[t]
  \centering
  \includegraphics[width=0.99\textwidth]{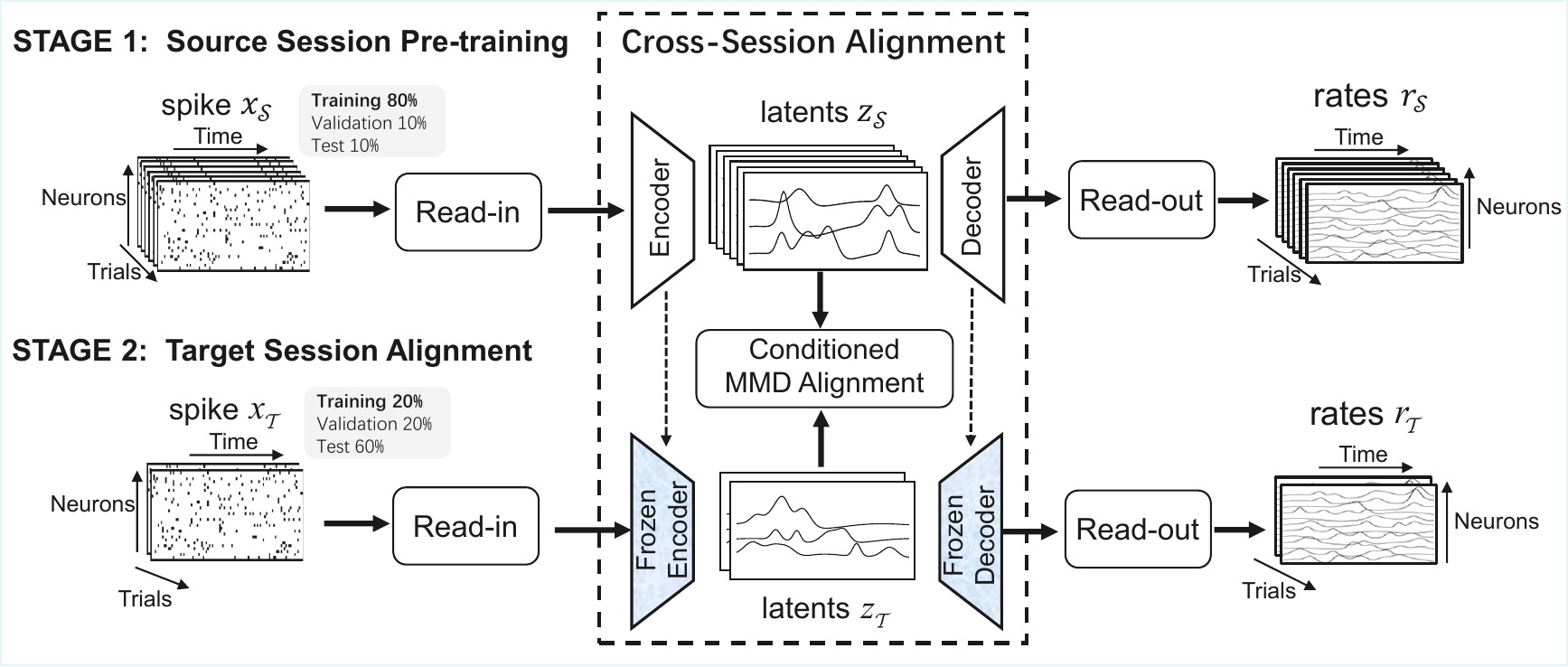} 
  \caption{Architecture of the proposed model. The model is composed of a shared autoencoder module and session-specific layers. Neural spiking data from a source session with sufficient trials are used to learn a latent representation via the encoder module. For target sessions with limited trials, alignment is performed separately for each task condition (movement direction in the center-out task) using multi-kernel MMD in latent space.
  }
  \label{Architecture}
\end{figure*}

\begin{itemize}
\item  $x_m^{\left( i \right)} \in {\mathbb{R}^{C_m \times T}}$ denotes binned neural spike counts (5 ms window length, non-overlapping) recorded simultaneously from $C_m$ channels over $T$ time bins,
\item $y_m^{\left( i \right)} \in {\mathbb{R}^{2 \times T}}$ denotes the 2-dimensional continuous kinematics (e.g., position or speed) over the same $T$ time bins as $x^{\left( i,m \right)}$,
\item $l_m^{\left( i \right)} \in \left\{ {1,2,...,D} \right\}$ denotes the behavioral condition ($D$ different movement directions) of the trial.
\end{itemize}

\subsection{Model Architecture}

Our model is built upon the autoencoder architecture introduced in LDNS. 
LDNS is a diffusion-based generative framework that employs an autoencoder to project spiking data into the latent space. In LDNS, both the autoencoder and the latent diffusion model use structured state-space (S4)\cite{gu2022efficiently} layers to model temporal dependencies. While LDNS further applies diffusion models on the latents to generate realistic neural spiking data, we focus solely on the autoencoder module for latent representation learning.
Given binned spike counts ${x_m^{\left( i \right)}}$ with $C_m$ units  from session $m$, the autoencoder encodes population activity into a $q$-dimensional ($q<C_m$) latent trajectory $z_m^{\left( i \right)} \in {\mathbb{R}^{q \times T}}$, and reconstructs smoothed firing rates $r_m^{\left( i \right)} \in {\mathbb{R}^{C_m \times T}}$that preserve the activity structure of the original neural population.

To handle data from multiple sessions with different numbers of recorded channels or neurons, 
TCLA integrates two components: (1) session-specific read-in and read-out layers, a pair of 1×1 convolutional (Conv1d) layers, and (2) a shared LDNS autoencoder module. This design allows neural recordings from different sessions to be processed within a unified latent space.

For each session, a session-specific read-in layer first projects the raw spiking activity into
an embedding with a fixed dimension that can be processed by the shared encoder. The shared encoder maps this embedding into a common latent space that is shared across all sessions, providing a unified representation of the neural population. Both source and target session data are projected into this shared latent space, where their representations can be directly compared and aligned. The shared decoder and session-specific read-out layers then map the latent representation back to reconstructed smoothed firing rates.

\begin{table}[t]
\centering
\small
\refstepcounter{table}
\label{tab:tcla_arch}

\begin{minipage}{1\linewidth}
\textbf{Table \thetable:} TCLA architecture hyperparameters.
The same setting was used for all datasets.
\textit{Decoder blocks = 0} means that no additional decoder block was used in the decoder stack. The decoder instead consisted of a GELU activation followed by the final output projection layer.
\end{minipage}

\vspace{0.4em}

\setlength{\tabcolsep}{5pt}
\begin{tabular}{ll}
\toprule
\textbf{Hyperparameter} & \textbf{Value} \\
\midrule
Shared encoder input dim & 30 \\
Hidden channel dim & 256 \\
Latent dimension & 16 \\
Encoder blocks & 4 \\
Decoder blocks & 0 \\
Linear layers per block & 2 \\
\bottomrule
\end{tabular}
\end{table}

\subsection{Two-Stage Training Framework}
The overall process of TCLA involves two stages.
\subsubsection{Source Session Pre-training}
In the first stage, both the shared autoencoder and the session-specific layers of the source session are optimized jointly. The objective is to learn a latent representation that captures the low-dimensional structure in neural population activity within the source session. Following LDNS, we optimize the autoencoder for spike reconstruction with latent regularization by minimizing the Poisson negative log-likelihood of the source-session spike counts $x_{\mathcal{S}}^{(i)}$ given the inferred firing rates $r_{\mathcal{S}}^{(i)}$. Additional regularization terms are applied to constrain the scale of the latent trajectories ($L_2$ regularization with hyperparameter $\beta_1$, selected from $[10^{-4},10^{-3}]$) 
and encourage temporal smoothness (regularization with hyperparameter $\beta_2$, selected from the range $[0.01,0.2]$) of $z_{\mathcal{S}}^{(i)}$:

\begin{equation*}
{\mathcal{L}_{\text{r}}} = {\beta _1}{\left\| {{z_{\mathcal{S}}^{(i)}}} \right\|^2} + {\beta _2}\sum\limits_{w = 1}^W {\sum\limits_{t = w + 1}^T {\frac{{{{\left\| {{z_{\mathcal{S}}^{(i)}}(t) - {z_{\mathcal{S}}^{(i)}}(t - w)} \right\|}^2}}}{{ {1 + w}}}} }.
\end{equation*}
Here, $W$ denotes the maximum window size used to enforce smoothness across neighboring latent time bins of $z_{\mathcal{S}}^{(i)}$.
The total loss in Stage 1 is defined as follows:


\begin{equation*}
{\mathcal{L}_{1}} = {E_{{x_{\mathcal{S}}} \sim U}}\left[\sum\limits_{i = 1}^n{( {\underbrace {r_{\cal S}^{(i)} - x_{\cal S}^{(i)}\ln r_{\cal S}^{(i)}}_{\mathrm{Poisson\;NLL}} + {{\mathcal L}_\text{r}}} )}\right].
\end{equation*}





\subsubsection{Target Session Alignment}
In the second stage, the model is adapted to a target session. To emulate conditions where limited data are available, only a small portion of data from the target session is used for training, while the majority is reserved for testing. During this stage, parameters of the shared autoencoder module are frozen, and only the session-specific layers of the target session are trained.
The objective is to simultaneously reconstruct the target spikes and align the conditional latent distributions with those of the source session.
Latent trajectories are grouped according to movement direction, and alignment is performed separately for each condition using multi-kernel Maximum Mean Discrepancy (multi-kernel MMD) \cite{gretton2012kernel} by minimizing the alignment loss ${\mathcal{L}}_{\text{MMD}}$:

\begin{equation*}
\mathcal{L}_{\mathrm{MMD}}
= \sum_{d=1}^{D}\!\Big[
k(z_\mathcal{S}^{(d)},z_\mathcal{S}^{(d)})
+ k(z_\mathcal{T}^{(d)},z_\mathcal{T}^{(d)})
- 2k(z_\mathcal{S}^{(d)},z_\mathcal{T}^{(d)})
\Big],
\end{equation*}
where $z_{\cal S}^{\left( d \right)}$ denotes the set of latents derived from trials in the source session corresponding to movement direction $d$, and $z_{\cal T}^{\left( d \right)}$ denotes the matching set of latents in the target session. The kernel function $k\left( { \cdot , \cdot } \right)$ is implemented as a Gaussian kernel,
\begin{equation*}
k\left( {A,B} \right) = \frac{1}{{\left| A \right|\left| B \right|}}\sum\limits_{j = 1}^J {\sum\limits_{{z_A} \in A} {\sum\limits_{{z_B} \in B} {\exp \left( { - \frac{{{{\left\| {{z_A} - {z_B}} \right\|}^2}}}{{{\sigma _j}}}} \right)} } }, 
\end{equation*}
where $\left|  \cdot  \right|$ represents the number of samples contained in the corresponding set. $\left\{ {{\sigma _j}} \right\}_{j = 1}^J $ denotes a set of bandwidths centered around the average pairwise distance across samples, with a scaling factor $K$ that controls the geometric progression of the bandwidths:
\begin{equation*}
\sigma_j
= \frac{K^{\, j - {\left\lfloor {\frac{J}{2}} \right\rfloor }}}{|A\cup B|(|A\cup B|-1)}
\sum_{\substack{z,z'\in A\cup B\\ z\neq z'}}
\|z-z'\|^2 .
\end{equation*}
Given spike counts $x_{\mathcal{T}}^{(i)}$ from a target session, this design results in the combined loss of Stage 2:

\begin{equation*}
{\mathcal{L}_{2}} = {\mathbb{E}_{{x_{\mathcal{T}}} \sim U}}\left[ {\sum\limits_{i = 1}^n {\underbrace {(r_{\cal T}^{(i)} - x_{\cal T}^{(i)}\ln r_{\cal T}^{(i)})}_{\mathrm{Poisson\;NLL}}}  + {\beta _3}{{\cal L}_{{\rm{MMD}}}}} \right],
\end{equation*}

where $r_{\mathcal{T}}^{(i)}$ denotes the inferred firing rates of the target session.
$\beta_3$ is a hyperparameter balancing reconstruction and alignment, selected from the range $[1,10]$.

In both training stages, we use coordinated dropout~\cite{keshtkaran2019enabling} by randomly masking input time bins and computing the loss only at the masked locations. This method prevents the autoencoder from simply memorizing observed spiking activity and encourages the model to infer smooth firing rates. Model parameters are optimized using the AdamW optimizer\cite{loshchilov2018decoupled}. The learning rate is scheduled using a cosine schedule with a linear warmup phase.





\section{EXPERIMENTS}
\subsection{Datasets}
\begin{figure}[htbp]
\centerline{\includegraphics[scale=0.8]{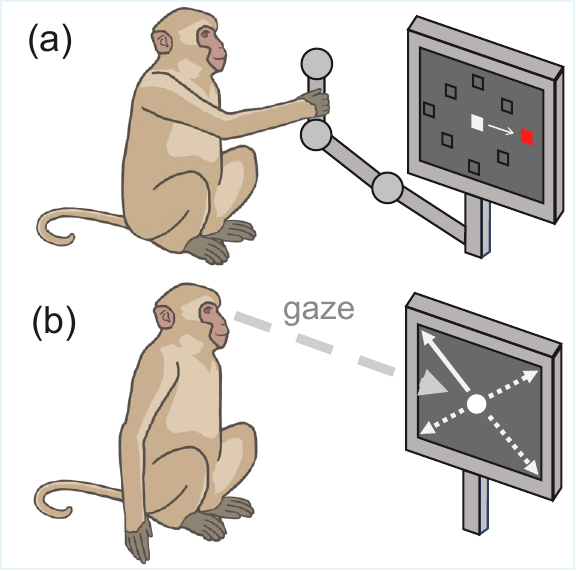}}
\caption{Experimental Paradigms. (a) Eight-direction motor center-out task. Two monkeys perform reaching tasks while neural activity is recorded from M1.
(b) Four-direction oculomotor center-out task. A monkey performs gaze-tracking tasks while neural activity is recorded from FEF and MT.
}
\label{experiment}
\end{figure}

We evaluated our framework on three neural spiking datasets collected from non-human primates performing center-out tasks, covering both arm movements and eye movements, as shown in Fig. \ref{experiment}.

\textbf{Motor center-out reaching datasets.}
We used two motor center-out datasets reported in~\cite{ma2023using}, collected from two monkeys (denoted as \textit{MOTORCO}\textsubscript{1} and \textit{MOTORCO}\textsubscript{2}) performing a center-out reaching task. 
In each trial, the monkeys controlled a cursor by moving a planar manipulandum with their arm, from a central starting position to one of eight peripheral targets.
Neural spiking activity was recorded from the Primary Motor Cortex (M1) using a chronically implanted 96-channel Utah array, with behavioral data recorded simultaneously. \textit{MOTORCO}\textsubscript{1} contains 12 recording sessions and \textit{MOTORCO}\textsubscript{2} contains 11 sessions. In both datasets, each session contains approximately 200 trials, with neural activity recorded from 91 to 96 channels per session. Each trial consists of 210 time bins.

\textbf{Oculomotor center-out gaze dataset.}
We also used an oculomotor center-out dataset (denoted as \textit{OCULOCO}) reported in~\cite{noneman2024decoding}. 
In this task, the monkey maintained gaze on a moving target as it moved from the center of the screen to one of four possible peripheral locations.
Spiking activity was recorded simultaneously from the Frontal Eye Fields (FEF) and Medial Temporal (MT) area, together with the corresponding behavioral measurements.
All procedures were approved by the Institutional Animal Care and Use Committee of Duke University.
\textit{OCULOCO} contains 6 recording sessions, each with approximately 2000 trials and 42 to 69 recorded neurons per session.  Each trial consists of 320 time bins.

\begin{figure*}[t]
  \centering
  \includegraphics[width=\textwidth]{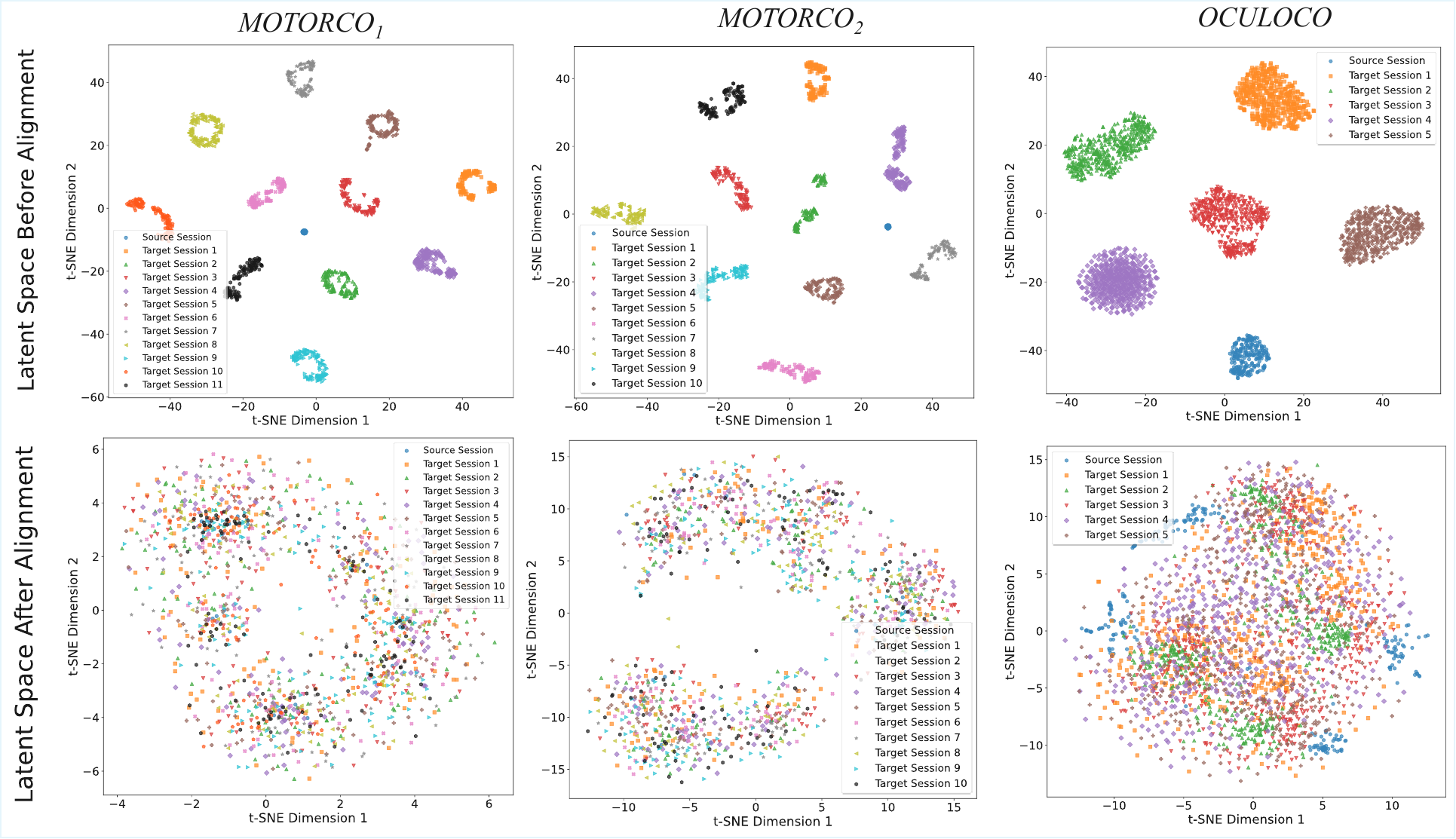}
  \caption{t-SNE visualization of the latent representations from the source and target sessions. Each column corresponds to one dataset (\textit{MOTORCO}\textsubscript{1}, \textit{MOTORCO}\textsubscript{2}, or \textit{OCULOCO}). The top row shows the latent representations learned by LDNSws, where the autoencoder is trained independently for each session without cross-session latent alignment. 
  The bottom row shows the latent representations learned by TCLA. The latent manifolds of the target sessions become more cohesive and better aligned with that of the source session.
  }
  \label{t-sne}
\end{figure*}

\subsection{Experimental Settings}
For each dataset, we designate one session as the source session and treat all remaining sessions as target sessions.
The source session is divided into training, validation, and test sets in an 8:1:1 ratio and used to pre-train the shared autoencoder.
To emulate a limited-data setting, we divide each target session into training, validation, and test sets in a ratio of 1:1:3.
All results are reported as the average over five independent runs for each target session.

\begin{table}[t]
\centering
\small
\refstepcounter{table}
\label{tab:hyperparams_short}

\begin{minipage}{1\linewidth}
\textbf{Table \thetable:} The hyperparameter settings used in Stage 1, Stage 2, and the downstream LSTM decoder training.
$\cdot$/$\cdot$/$\cdot$ denote the hyperparameter settings for 
\textit{MOTORCO}\textsubscript{1},
\textit{MOTORCO}\textsubscript{2}, and \textit{OCULOCO}, respectively. A single value indicates that the same setting was used for all datasets.
\end{minipage}

\vspace{0.4em}

\setlength{\tabcolsep}{5pt}
\begin{tabular}{ll}
\toprule
\textbf{Hyperparameter} & \textbf{Value} \\
\midrule
\multicolumn{2}{l}{\textbf{Stage 1}} \\
Learning rate & $10^{-3}$\\
batch size &  64 \\
Epochs & 400/300/300 \\
Warmup epochs & 100 \\
Smoothness window $W$ & 5 \\
$\beta_1$ & $10^{-4}$/$10^{-4}$/$10^{-3}$ \\
$\beta_2$ & 0.01/0.01/0.2 \\
Mask probability & 0.5 \\
Grad clip & 1.0 \\
\midrule
\multicolumn{2}{l}{\textbf{Stage 2}} \\
Learning rate & $10^{-3}$ \\
Batch size & 32 \\
Epochs & 400 \\
Warmup epochs & 10 \\
$\beta_3$ & 1/1/10 \\
Kernel scaling factor $K$ & 2.0 \\
Number of Gaussian kernels $J$ & 5\\
Mask probability & 0.5 \\
Grad clip & 1.0 \\
\midrule
\multicolumn{2}{l}{\textbf{LSTM decoder}} \\
Hidden units & 256 \\
Dropout & 0.1 \\
Epochs & 25 \\
Batch size & 64 \\
Learning rate & $10^{-3}$ \\
Bins before & 6 \\
Bins current & 1\\
Bins after & 0 \\
\bottomrule
\end{tabular}
\end{table}

\section{RESULTS}
We compared TCLA with two baseline methods:
\begin{itemize}
\item AutoLFADS\cite{pandarinath2018inferring}, \cite{keshtkaran2022large}, a widely used latent variable model that infers firing rates from spiking activity and is trained independently on each target session.
\item LDNS-within-session (LDNSws), which adopts the same autoencoder architecture from LDNS and uses session-specific layers as TCLA, but is trained solely on target-session data without cross-session alignment.
\end{itemize}

Decoding performance is quantified using the coefficient of determination (${R^2}$) between the predicted and the ground-truth behavioral variables. 
To obtain a robust estimate of performance, we report the bootstrap mean and 95\% confidence intervals (CI) of $R^2$ across target sessions (10,000 resamples).
For each target session, a separate downstream Long Short-Term Memory (LSTM) decoder is trained on the inferred firing rates to predict the corresponding behavioral variables. Each LSTM decoder is optimized using the RMSprop optimizer. Both the inferred firing rates and the behavioral variables are resampled to a bin size of 50 ms before LSTM decoding.





\subsection{Conditionally Aligned Latent Manifolds}
We first assess whether TCLA effectively aligns neural representations across recording sessions.
To visualize the structure of the learned latent space, we apply t-distributed Stochastic Neighbor Embedding (t-SNE)\cite{maaten2008visualizing} to latent trajectories inferred from the test sets of both the source session and each target session.
Latent representations for all three datasets are shown in Fig. \ref{t-sne}.

When the autoencoder is trained independently on each target session without cross-session alignment (LDNSws), latent representations 
form separated clusters, indicating substantial inter-session mismatch.
In contrast, TCLA projects neural activity from target sessions into a latent space that is aligned with the source session, resulting in substantially improved overlap and reduced cross-session dispersion.

Together, these results suggest that TCLA effectively maps spiking activity from target sessions into a shared latent space learned from the source session, thereby promoting cross-session consistency in neural representations.

\subsection{Decoding Performance}

\begin{figure*}[t]
  \centering
  \includegraphics[width=\textwidth]{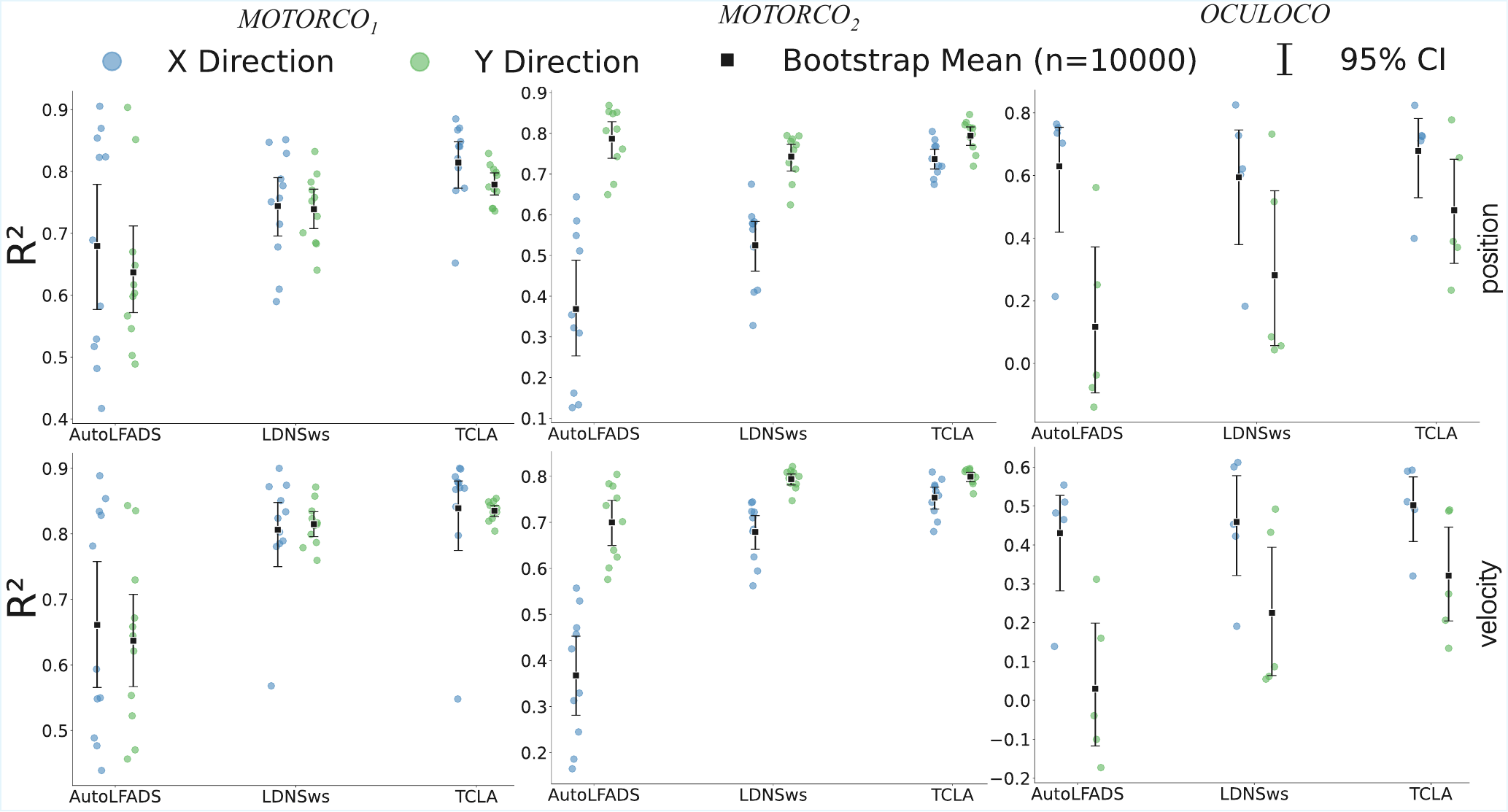} 
  \caption{Comparison of decoding performance across methods and datasets. Each column represents a different dataset. The top row shows $R^2$ for position decoding, and the bottom row shows the results for velocity decoding.
  For each method and kinematic prediction, performance is shown separately for the $x$ (blue) and $y$ (green) coordinates. Each dot represents the mean $R^2$ of a target session.
  The solid black square marker indicates the bootstrap mean, and the error bar represents the 95\% CI across all target sessions.}
  \label{R2}
\end{figure*}
We next quantify whether this cross-session alignment translates into improved downstream decoding performance.

As shown in Fig. \ref{R2}, 
TCLA consistently improves decoding accuracy relative to both baselines across all three datasets, for both position and velocity decoding.
For the two motor datasets (\textit{MOTORCO}\textsubscript{1} and \textit{MOTORCO}\textsubscript{2}), TCLA yields significant gains
in decoding the $x$ coordinate across sessions for both position and velocity variables (mean improvement from 0.033 to 0.386 in $R^2$, $P<10^{-4}$ in these cases, Wilcoxon signed-rank test). For \textit{OCULOCO}, decoding performance improves significantly for the $y$ coordinate (mean improvement from 0.093 to 0.374 in $R^2$, $P<10^{-5}$ in these cases, Wilcoxon signed-rank test).

The LDNSws baseline, which shares the same autoencoder backbone as TCLA but is trained only on the limited target-session data, achieves a higher average $R^2$ than AutoLFADS in most settings, yet remains consistently inferior to TCLA. 
This comparison indicates that the performance gains are not solely attributable to the autoencoder architecture, but instead arise from effective knowledge transfer enabled by the proposed task-conditioned cross-session alignment.

\section{DISCUSSION}

Across three neural spiking datasets involving both motor and oculomotor center-out paradigms, TCLA demonstrates robust improvements over AutoLFADS and LDNSws. Importantly, the consistent improvement over LDNSws highlights that the primary benefit of TCLA stems from cross-session transfer enabled by task-conditioned alignment, rather than from architectural capacity alone.

Across all datasets, the performance gains achieved by TCLA are closely related to the baseline decoding performance of each coordinate. In both \textit{MOTORCO}\textsubscript{1} and \textit{MOTORCO}\textsubscript{2}, improvements are more pronounced along the $x$ coordinate, where the baseline decoding performance is relatively weaker. In contrast, for \textit{OCULOCO}, significant improvements are observed along the $y$ coordinate, corresponding to the coordinate with lower baseline decoding accuracy. 
This result suggests that the relative effectiveness of knowledge transfer depends strongly on the baseline performance of the target session: coordinates with weaker initial decoding performance tend to benefit more from knowledge transfer, while those with stronger baselines exhibit smaller relative gains. This behavior is consistent with observations in transfer learning that improvements are often more pronounced when the target domain exhibits lower baseline performance \cite{zhuang2020comprehensive}.

Beyond improved $R^2$, the representation visualizations (Fig. \ref{t-sne}) suggest that TCLA promotes a more session-consistent latent organization, reducing session-specific distortions that can hinder decoding. This indicates that conditioning alignment on behavioral context helps preserve structure that is meaningful for downstream prediction. Notably, the observed gains arise even though the behavioral decoder is trained separately for each session, implying that TCLA improves the quality and comparability of inferred latent trajectories in a way that facilitates decoder training under limited target-session data.

Taken together, these findings support TCLA as a practical approach for leveraging previously collected data to improve cross-session decoding performance when only limited target-session data are available.


\section{CONCLUSION}
Training accurate neural decoders can be difficult when only limited data are available from a recording session. In this work, we proposed TCLA, a task-conditioned latent alignment framework that transfers low-dimensional neural representations learned from a source session with sufficient data to target sessions with limited data. Across three non-human primate center-out datasets spanning both arm-movement and oculomotor behaviors, TCLA consistently improved decoding accuracy relative to AutoLFADS and an architecture-matched within-session baseline, demonstrating the value of task-conditioned cross-session transfer beyond architectural effects alone. Overall, TCLA provides a practical strategy for improving cross-session generalization under realistic data constraints.
One limitation is that TCLA relies on task-condition labels to perform conditional alignment across sessions. In scenarios where task labels are ambiguous or unavailable, additional strategies may be required.
Future work will investigate settings in which a decoder trained on the source session is held fixed and applied directly to target sessions via aligned representations, reducing the need for per-session decoder retraining. Exploring real-time deployment and extending the approach to more complex behaviors may further strengthen the applicability of TCLA to long-term invasive BCI systems.

\bibliography{ref}

\end{document}